\pdfoutput=1
\documentclass[11pt]{article}

\usepackage[]{ACL2023}
\usepackage{times}
\usepackage{latexsym}
\usepackage{comment}
\usepackage{graphicx}
\usepackage{subcaption}
\usepackage{amsmath}
\usepackage{booktabs}
\usepackage[T1]{fontenc}
\usepackage[utf8]{inputenc}
\usepackage{microtype}
\usepackage{inconsolata}

\title{Unveiling the Role of Pretraining in Direct Speech Translation}

\author{Belen Alastruey$^{1,2}$, Gerard I. Gállego$^2$, Marta R. Costa-jussà$^1$ \\
           FAIR, Meta$^1$\\
           TALP Research Center, Universitat Politècnica de Catalunya, Barcelona$^2$ \\
         \texttt{alastruey@meta.com,gerard.i.gallego@upc.edu,costajussa@meta.com}
         }

\begin{document}
\setlength{\abovedisplayskip}{1pt}

\maketitle
\begin{abstract}
Direct speech-to-text translation systems encounter an important drawback in data scarcity. A common solution consists on pretraining the encoder on automatic speech recognition, hence losing efficiency in the training process. In this study, we compare the training dynamics of a system using a pretrained encoder, the conventional approach, and one trained from scratch. We observe that, throughout the training, the randomly initialized model struggles to incorporate information from the speech inputs for its predictions. Hence, we hypothesize that this issue stems from the difficulty of effectively training an encoder for direct speech translation. While a model trained from scratch needs to learn acoustic and semantic modeling simultaneously, a pretrained one can just focus on the latter.  Based on these findings, we propose a subtle change in the decoder cross-attention to integrate source information from earlier steps in training. We show that with this change, the model trained from scratch can achieve comparable performance to the pretrained one, while reducing the training time.

\end{abstract}

\section{Introduction}
\label{sec:intro}
In recent years, extensive research has been done in the field of speech-to-text translation (ST). These models have transitioned from cascaded systems to direct ones \cite{iwslt-2023-international}. While this shift helps mitigate error propagation, it introduces other challenges such as scarcity of training data and the need for the model to tackle translation and speech recognition simultaneously. To bypass these issues, a common approach to train direct ST systems involves pretraining the encoder on the Automatic Speech Recognition (ASR) task \citep{berard2018end}. This enables the encoder to learn acoustic modeling in the source language by leveraging ASR data, and the model can focus on semantic modeling during the ST training. 

Various studies have been conducted on pretraining for ST. \citet{bansal-etal-2019-pre}, introduced a method to enhance ST performance for low-resource source languages by utilizing ASR pretraining from a high-resource language. \citet{alinejad-sarkar-2020-effectively} enhanced the performance of an ST system by pretraining both the encoder and the decoder on ASR and MT respectively. \citet{wang-etal-2020-curriculum} and \citet{le2023pretraining} proposed variations of ASR pretraining that yielded superior results. 

However, pretraining has some drawbacks too. It has additional data requirements, which can be a problem particularly in languages that don't have a written form and hence no ASR data. Furthermore, it  complicates the training pipeline and worsens the efficiency of the overall training process. 

Recent studies have already questioned the pretraining approach, \citet{zhang2022revisiting}, demonstrating that similar results can be achieved under certain conditions without the need for pretraining. However, the authors show that many strategies need to be simultaneously used to achieve this, such as an exhaustive hyperparameter tunning, CTC-based regularization and their proposed parameterized distance penalty.

Complementing previous interpretability works in ST \cite{xu-etal-2021-stacked, alastruey-etal-2022-locality}, in this study, we conduct the first-ever analysis of the training dynamics of a ST system, and based on its results, we propose a subtle modification in the Transformer \cite{vaswani2017attention} architecture to bypass the pretraining stage. 

First, we compare the training dynamics of a conventional system that uses a pretrained encoder with one trained from scratch\footnote{The pretraining is done on the same amount of training data than the ST training.}. Through this analysis, we observe significant disparities in their behaviors. Particularly, we note that when making predictions, the model trained from scratch delays the utilization of information extracted by the encoder until a later stage of training. 

We hypothesize that this delay occurs due to the complexity of the acoustic modeling task, that in this setting needs to be learned together with the semantic modeling. Hence, it takes a significant amount of updates to sufficiently train the encoder so that it can extract meaningful information. Consequently, the model ignores the encoder outputs and focuses on training the decoder for language modeling. Once the encoder can extract valuable representations, the model has already converged towards language modeling and struggles to rely on the information obtained by the encoder.

Secondly, we believe that by forcing the model to utilize encoder outputs earlier in the training process, the model would not converge towards language modeling and the encoder would be trained more rapidly, leading to higher-quality representations in its outputs. Through a modification in the residual connection in the decoder cross-attention mechanism, we force the model trained from scratch to integrate source information from earlier training steps, and we observe a comparable performance than in the pretrained one.

Overall, the main contributions of our work are: (1) the first study of training dynamics in ST, that unveils the role of the pretraining, and (2) a modification in the Transformer architecture to bypass the pretraining stage.

\section{Related Work}

\subsection{Interpretability of Transformer Models}
\label{sec:interpretability_method}
Our research aims to quantify the source of information used for making predictions in these models, specifically whether it originates from the source (encoder input) or the target prefix (previously predicted words serving as decoder inputs). To achieve this, we employ the ALTI+ interpretability method \cite{ferrando-etal-2022-towards}. 

ALTI+ employs a strategy to rewrite attention blocks, as introduced by \citet{kobayashi-etal-2020-attention}, along with the contribution definition provided by \citet{ferrando-etal-2022-measuring} and a variation of rollout inspired by \citet{abnar-zuidema-2020-quantifying}. By utilizing ALTI+, we determine the extent to which each input token in the source and target prefix contribute to the prediction of a token. Furthermore, by summing the individual contribution of each token in the encoder source, the authors obtain a unique score referred to as source contribution, that we use to study training dynamics on ST.

\subsection{Training Dynamics on Machine Translation}
\label{sec:lena}
Previous work has been done to understand how Transformers learn in the task of Machine Translation on text. \citet{voita-etal-2021-language} analyse how the source contribution varies along the training using Layerwise Relevance Propagation method to track source and target contribution, and describe three different training phases.
\paragraph{Target-side language modeling:} The beginning of training is devoted to target-side language modeling. The total contribution of the source substantially decreases. This means that in the trade-off between information coming from the source and the target prefix, the model gives more and more priority to the prefix.
\paragraph{Learning how to use source:} In the second stage, the source influence increases quickly. This means that, opposite to the first stage, in the trade-off between information coming from the source and the target prefix, the model progressively gives more and more importance to source information.
\paragraph{Refining translations:} In the last stage, the source contribution remains constant. By analysing other metrics the authors see that the model is learning to refine some translations. The model learns to align better, and is able to generate more natural translations instead of word-to-word ones.

\begin{figure*}
    \centering
    \includegraphics[width=0.9\linewidth]{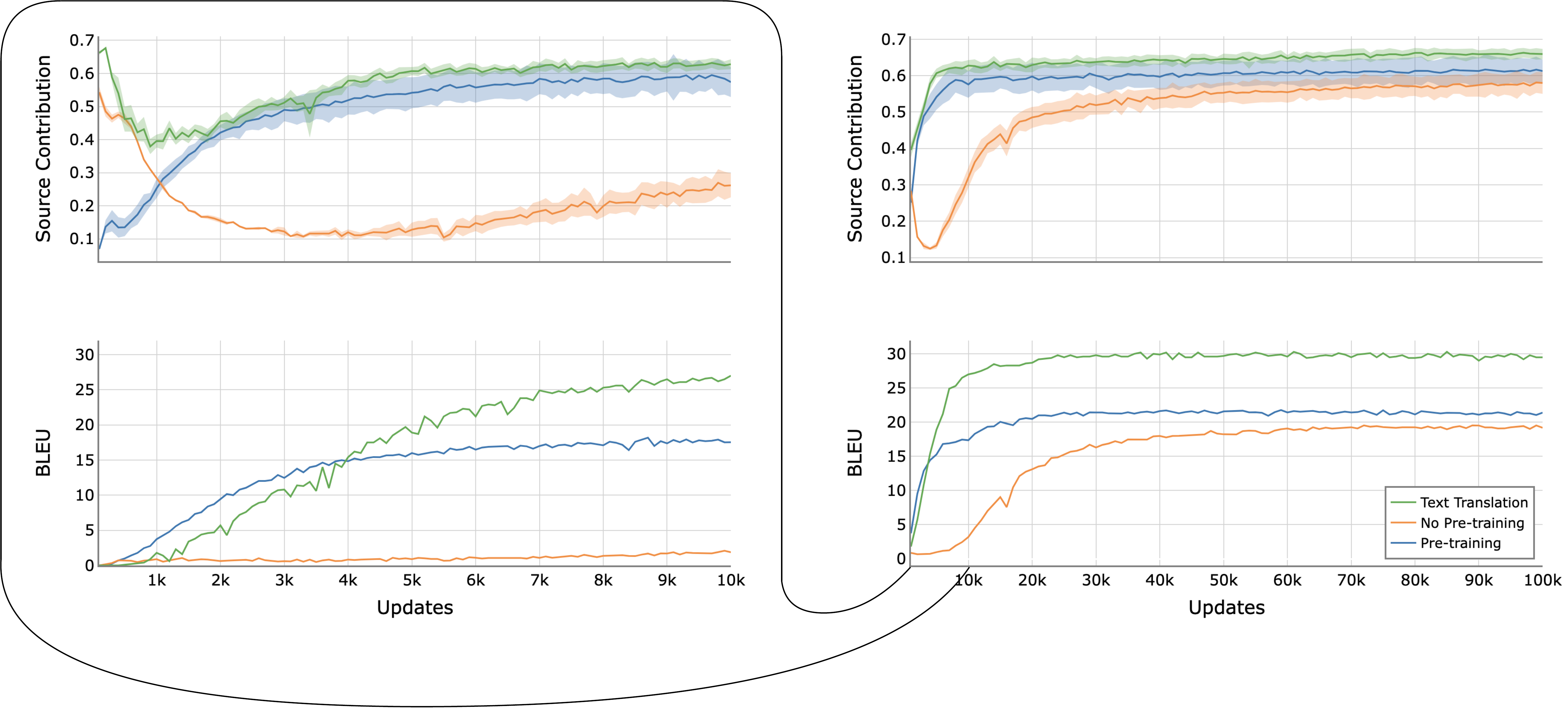}
    \caption{Source contribution ($\pm$std) and BLEU along the full training (right) and along the first 10k updates (left).}
    \label{fig:contribs}
\end{figure*}

\section{Training Dynamics in Speech Translation}

In this section, we analyse how much two training strategies on a Transformer-based system\footnote{We train the small S2T-Transformer architecture from Fairseq (\url{https://github.com/facebookresearch/fairseq}).} rely on the speech source when making predictions along a ST training. In particular, we study: (1) a standard ST system,  consisting on an ASR pretraining, followed by a re-initialization of the decoder and a training on ST data, and (2) a system  that only performs a ST training on a randomly initialized model.

To measure the amount of input information used by the model to generate a prediction we use the source contribution defined by ALTI+, covered in Section \ref{sec:interpretability_method}. 

To generalize this score to a sentence-wise score, we average the source contribution used for the prediction of each token in a sentence. Finally, to obtain a score over a test set, we average again the score obtained by every sentence in the set.

Given that we use a different source contribution measure than \citet{voita-etal-2021-language} in their work described in Section \ref{sec:lena}, we decide to train an additional MT model, to confirm that the three stages described in their work still happen on our setting.

For all our analysis, we store a checkpoint every 1k updates during the full training, and every 100 updates during the first 10k\footnote{Setup details of the experiments are in Appendix \ref{appx:exp_setup}.}. For each of the checkpoints, we evaluate the model computing BLEU and source contribution scores on English-German MuST-C dataset \cite{Cattoni2021} \footnote{We use transcripts and text translations for the MT model.}.

\subsection{Results Analysis}
\label{sec:training_dynamics_analysis}

In Figure \ref{fig:contribs}, we see the obtained results. When focusing on the first 10k updates we first observe that the three stages described in Section \ref{sec:lena} still happen in the text translation model. However, when analysing both ST variants, we observe different behaviours. 

In the standard setting with a pretrained encoder, we observe a two-stage process. This model skips the first stage described in Section \ref{sec:lena}, and rapidly integrates source data from the beginning of training. This phenomenon is coherent, as the encoder has been pretrained, resulting in high-quality representations that are immediately beneficial for the decoder during prediction tasks. As in the case of text translation, the last stage starts after the first around 6k updates.

Instead, the ST model trained from scratch undergoes the same three-stage process than text translation. However, each stage appears to require significantly more time compared to text translation. Specifically, the model does not achieve a stable level of source contribution until after approximately 30k updates, whereas the other two models achieve this stability after only 6k updates.

We hypothesize this happens due to the difficulty of training the encoder for the task of ST from scratch. Unlike an encoder in text translation, which solely requires semantic modeling, a ST encoder learns both acoustic and semantic modeling. This dual requirement makes the training process for an ST encoder more time-consuming than that of a text translation model.   Consequently, the model tends to overlook the encoder during the early stages of training, focusing instead on language modeling. 

Overall, we believe that the initial stage outlined in Section \ref{sec:lena} is not a result of the need to learn language modeling. Rather, it's a strategy to bypass the encoder information until the encoder is adequately trained. This process is quick in text translation, non-existent when using a pre-trained encoder in ST, and lengthy when training an ST system from scratch.

Moreover, we think that by the time the ST encoder trained from scratch becomes capable of extracting relevant information, the model has already converged towards relying on language modeling. As a result, it never reaches the level of contribution achieved by the pretrained model (as shown in Figure \ref{fig:contribs}), leading to inferior performance.

\section{Training ST from Scratch}
\label{sec:skip_pretraining}

Building on our previous analysis,  we hypothesize that forcing a Speech Translation model trained from scratch to utilize source information from the start could enhance the training process. If the model is required to use the encoder's representations, regardless of their quality, poor representations will negatively impact the model's overall performance. This, in turn, will cause a faster training of the encoder to extract better representations.

In particular, considering the results in Figure \ref{fig:contribs}, we observe that both the text translation model and the pretrained speech translation model achieve a stable source contribution of approximately 65\%. Hence, we consider this proportion to be optimal and aim to enforce it in the speech translation model trained from scratch.

To test our hypothesis we propose a subtle architecture modification, that forces the Transformer to use source information along the full training. Our modification focuses on the cross-attention layer of the decoder, which is the step where source and target information is aggregated, with source information coming from the attention block and target-prefix details coming from the residual flow.

\begin{figure}
    \centering
    \includegraphics[width=1\linewidth]{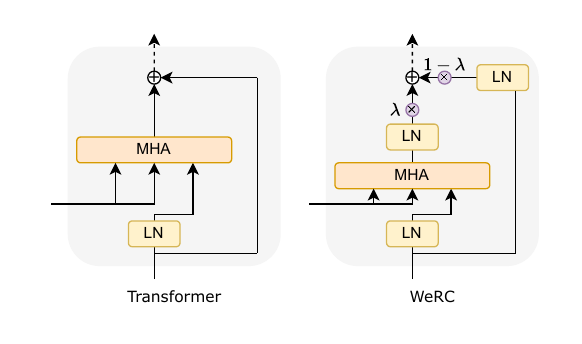}
    \caption{Standard S2T-Transformer cross-attention layer (left) and proposed WeRC (right).}
    \label{fig:werc}
\end{figure}

\subsection{WeRC: Weighted Residual Connection}

We modify the residual connection after each cross-attention block in the decoder. This sum aggregates the output of the cross-attention block ($x_{attn}$) and the residual stream($x_{res}$). Our goal is to increase the information flow coming from the source, so we scale these two components giving a higher weight to the output of the cross-attention (Eq. \ref{eq:werc_1}). Specifically, we aim to approximately match the proportion of source contribution found in Section \ref{sec:training_dynamics_analysis}, hence we set $\lambda = 0.65$.

\begin{equation}
\label{eq:werc_1}
x_{out} = \lambda \cdot x_{attn} + (1 - \lambda) \cdot x_{res}
\end{equation}

However, a potential issue of this approach is that the cross-attention block could converge towards producing small-norm vectors, so that they would still have a small contribution regardless of the weighting. To solve this potential issue, we normalize each term of the summation (Eq. \ref{eq:werc_2}), by adding layer normalization layers \cite{ba2016layer}. This ensures both tensors have the same norm before the weighting. Therefore they will contribute to the sum with the target proportion.\footnote{Note that we remove the learnable parameters from layer normalization to avoid any scaling that could affect the predefined weights.}

\begin{equation}
\label{eq:werc_2}
x_{out} = \lambda \cdot LN(x_{attn}) + (1 - \lambda) \cdot LN(x_{res})
\end{equation}

Results in Table \ref{tab:st_no_pt} show that our model, which incorporates WeRC and is trained from scratch, outperforms the baseline by +1.3 BLEU points. Additionally, it nearly achieves the same performance as the model with pretraining, while reducing the training time by skipping the pretraining stage. Additionally, we extend this experiment to En-Es and En-Fr MuST-C sets and obtain analogous results.

\begin{table}
\centering
\setlength{\tabcolsep}{3.5pt} 
\scalebox{0.85}{ %765
\begin{tabular}{lcccc}
\toprule
\textbf{Model} & \textbf{Pretrained} & \textbf{En-De} & \textbf{En-Es} & \textbf{En-Fr}\\
\toprule
Baseline & Yes & 22.4 & 27.3 & 32.3 \\ %21.8
Baseline & No &  20.4& 26.3 & 30.9 \\ %20.2
\midrule
WeRC & No &  22.2 & 27.2 & 32.4 \\ %21.5
\toprule

WeRC w/o norm & No & 21.8 & - & -\\ %20.8
WeRC w/o weights & No & 21.6 & - & -\\ %20.8
\bottomrule
\end{tabular}
}
\caption{BLEU ($\uparrow$) on MuST-C test set averaging the best 10 checkpoints.}
\label{tab:st_no_pt}
\end{table}

\subsection{Ablation Study}
We perform an ablation study on the usefulness of the weighted sum and the layer normalization individually. In Table \ref{tab:st_no_pt} we observe that both strategies achieve a better performance than the baseline trained from scratch, but they are still considerably behind WeRC and the pretrained baseline. In the case of the variant without normalization, we believe this is as a result of the trainable parameters in the attention block (as described earlier). In the case of the model without weights, we believe this happens because the model is forced to use a 50\% of source contribution, which is below the optimal (as observed in Figure \ref{fig:contribs}). Additional ablation studies regarding the use of WeRC on a MT and a pretrained ST models can be found in Appendix \ref{appx:ablation2}. %Furthermore, results on WeRC when applied to different languages available in the MuST-C dataset can be found in Appendix \ref{appx:languages}.

\section{Conclusions}
In this work, we present the first study on the training dynamics of direct ST systems, comparing a standard ST model with a pretrained encoder to one trained from scratch. The analysis shows that, without pretraining, the model struggles to incorporate information from the encoder's outputs when making predictions. As an explanation, we suggest the encoder needs more updates than in a text task until it can extract valuable representations of the input tokens. Once this is achieved, the model has already converged towards language modeling, hence failing to utilize the information extracted by the encoder effectively even in later steps. To address this issue, we propose a subtle modification to the transformer architecture that forces the model to incorporate source information throughout the whole training. By doing so, we achieve comparable performance in a model trained from scratch to one with pretraining, while reducing training time and data requirements.

\section*{Limitations}
While our study provides valuable insights into the training dynamics of direct ST systems and proposes a novel approach to improve the efficiency of the training process, our findings are based on a specific model, dataset and languages. We believe different results could be obtained in other settings, such as low resource speech translation.

Furthermore, our paper focuses on a classic and widely extended pretraining strategy. ASR and ST training sets correspond to the same dataset and have the same size, differing only in the language of the targets. We also don't use additional techniques such as CTC auxiliary loss. However, our goal in this work is not obtaining a new state-of-the-art ST training strategy but analysing and understanding a common training strategy using interpretability tools, and performing additional experiments to validate the hypothesis extracted from the analysis.

Finally, in our work we use the learning rate defined by \citet{Wang2020a} for ST finetuning also on the experiments trained from scratch. We acknowledge that the performance of experiments trained from scratch could be pushed further by tuning this hyperparameter. However, we wanted to keep experiments comparable for the training dynamics analysis, and hence we decided to use the same learning rate. Furthermore, this should not have an impact in the conclusions of the paper, given that our proposed modification (WeRC) is also trained from scratch and uses the same learning rate.

\bibliography{anthology,custom}
\bibliographystyle{acl_natbib}

\appendix

\section{Experimental Setup}
\label{appx:exp_setup}

\paragraph{ST Models Details:} In our experiments we use commonly used Fairseq\footnote{https://github.com/facebookresearch/fairseq} Transformer and S2T-Transformer architectures. In the case of speech, it consists of 12 encoder layers and 6 decoder layers. Both the encoder and the decoder use 4 attention heads, the embedding dimension is 256, and in the MLP blocks it is 2048. The decoder output dimension is 256, the same as the decoder embedding dimension. The model has layer normalization before its main blocks instead of after, and a dropout of 0.1 is used in both the attention weights and in the MLP activations. ReLU is used as the activation function for the MLP. Regarding text models we have 6 encoder and 6 decoder layers, no dropout, an embedding dimension of 512 and 8 attention heads (other settings remain the same than for speech).

\paragraph{Training Setup:} In the case of speech translation and speech recognition trainings, we follow the setup defined by \cite{Wang2020a}. We fix a maximum of 20000 tokens per batch. We use Adam optimizer \cite{kingma2017adam} and a learning rate of $1\cdot10^{-3}$ with an inverse square root scheduler. We apply a warm-up for the first 10000 updates and we clip the gradient to 10 to avoid exploding gradients. We use label smoothed Cross-entropy loss, with a smoothing factor of 0.1. The update frequency is set to 16, simulating the use of 16 GPUs. We train each model for a maximum of 100000 updates. In ST trainings we use a learning rate of $2\cdot10^{-3}$ while on speech recognition it is $1\cdot10^{-3}$, as done by \cite{Wang2020a}. 

In the text translation system, we again follow the setup defined by \cite{Wang2020a} for Machine Translation. It is similar than the speech translation one but the maximum number of tokens per batch is limited to 4096 and the number of warm updates is 4000. Gradient clipping is removed and learning rate is set to $5\cdot10^{-4}$.

\section{Results on Machine Translation and Speech Translation with Pretraining}
\label{appx:ablation2}
In this section, we aim to study the impact of using WeRC on the analysed MT system and on the ST training with a pretrained encoder. These settings achieve an optimal level of source contribution from the first updates of the training, so we hypothesize that WeRC might have a less noticeable impact than in the main study of this paper (ST from scratch).
In Table \ref{tab:machine_translation}\footnote{Note that this results are obtained evaluating on the best checkpoint without checkpoint averaging.} we see the obtained results. We observe that both settings maintain the same performance, which is consistent with our hypothesis.

\begin{table}
\centering
\begin{tabular}{lc}
\toprule
\textbf{Model} & \textbf{BLEU}\\
\toprule
Baseline MT & 31.6 \\
WeRC & 31.6 \\
\midrule
Baseline Pretrained ST & 21.8\\
WeRC & 21.8 \\
\bottomrule
\end{tabular}
\caption{WeRC performance on MT and pretrained ST.}
\label{tab:machine_translation}
\end{table}

\end{document}